\documentclass[a4paper,fleqn]{cas-dc}

\usepackage{duckuments}
\usepackage{titlesec}
\usepackage{amsmath} 
\usepackage{algorithm}
\usepackage{algpseudocode}
\usepackage{color}
\usepackage{tabularx}
\usepackage{booktabs}

\titleformat*{\section}{\bfseries\Large}
\usepackage{caption}
\usepackage{hyperref}
\hypersetup{colorlinks=true, citecolor=blue, linkcolor=blue, urlcolor=blue}

\def\tsc#1{\csdef{#1}{\textsc{\lowercase{#1}}\xspace}}
\tsc{WGM}
\tsc{QE}

\begin{document}
\let\WriteBookmarks\relax
\let\printorcid\relax
\def\floatpagepagefraction{1}
\def\textpagefraction{.001}

\shorttitle{}    

\shortauthors{}  

\title [mode = title]{ScalingGaussian: Enhancing 3D Content Creation with Generative Gaussian Splatting}  



%

\author[1]{Shen Chen}



\ead{y30220968@mail.ecust.edu.cn}







\credit{Writing – review \& editing, Writing – original draft, Software, Methodology, Investigation, Conceptualization, Resources, Validation, Data curation, Visualization}

\author[1]{Jiale Zhou}
\cormark[1]
\ead{zhou.jiale@ecust.edu.cn}
\credit{Writing – review \& editing, Writing – original draft, Software, Methodology, Conceptualization, Resources, Validation, Formal analysis, Data curation, Visualization}

\author[2]{Zhongyu Jiang}
\ead{zyjiang@uw.edu}
\credit{Writing – review \& editing, Validation}

\author[4]{Tianfang Zhang}
\ead{sparkcarleton@gmail.com}
\credit{Writing – review \& editing, Validation}

\author[5]{Zongkai Wu}
\ead{wuzongkai@fancy.tech}
\credit{Writing – review \& editing, Validation}

\author[2]{Jenq-Neng Hwang}
\ead{hwang@uw.edu}
\credit{Writing – review \& editing, Methodology}

\author[2,3]{Lei Li}
\cormark[1]

\ead{lilei@di.ku.dk}

\credit{Writing – review \& editing, Software, Methodology, Conceptualization, Resources, Validation, Visualization}

\affiliation[1]{organization={School of Information Science and Engineering, East China University of Science and Technology}, 
                city={Shanghai}, 
                postcode={200237}, 
                country={China}}
\affiliation[2]{organization={Department of Electrical \& Computer Engineering, University of Washington},
            city={Seattle},
            postcode={98195}, 
            state={WA},
            country={USA}}

\affiliation[3]{organization={Computer Science Department, University of Copenhagen},
            city={Copenhagen},
            postcode={2100}, 
            state={Capital Region},
            country={Denmark}}

\affiliation[4]{organization={Department of Automation, Tsinghua University}, 
                city={Beijing}, 
                postcode={100084}, 
                country={China}}

\affiliation[5]{organization={FancyTech}, 
                city={Hangzhou}, 
                postcode={310023}, 
                country={China}}

\cortext[1]{Corresponding author}



\begin{abstract}
The creation of high-quality 3D assets is paramount for applications in digital heritage preservation, entertainment, and robotics. Traditionally, this process necessitates skilled professionals and specialized software for the modeling, texturing, and rendering of 3D objects. 
However, the rising demand for 3D assets in gaming and virtual reality (VR) has led to the creation of accessible image-to-3D technologies, allowing non-professionals to produce 3D content and decreasing dependence on expert input.
Existing methods for 3D content generation struggle to simultaneously achieve detailed textures and strong geometric consistency.
We introduce a novel 3D content creation framework, ScalingGaussian, which combines 3D and 2D diffusion models to achieve detailed textures and geometric consistency in generated 3D assets. Initially, a 3D diffusion model generates point clouds, which are then densified through a process of selecting local regions, introducing Gaussian noise, followed by using local density-weighted selection. To refine the 3D Gaussians, we utilize a 2D diffusion model with Score Distillation Sampling (SDS) loss, guiding the 3D Gaussians to clone and split. 
Finally, the 3D Gaussians are converted into meshes, and the surface textures are optimized using Mean Square Error(MSE) and Gradient Profile Prior(GPP) losses.
Our method addresses the common issue of sparse point clouds in 3D diffusion, resulting in improved geometric structure and detailed textures. Experiments on image-to-3D tasks demonstrate that our approach efficiently generates high-quality 3D assets.

\end{abstract}




\maketitle
\begin{figure*}[t]
   \centering
   \includegraphics[width=\linewidth]{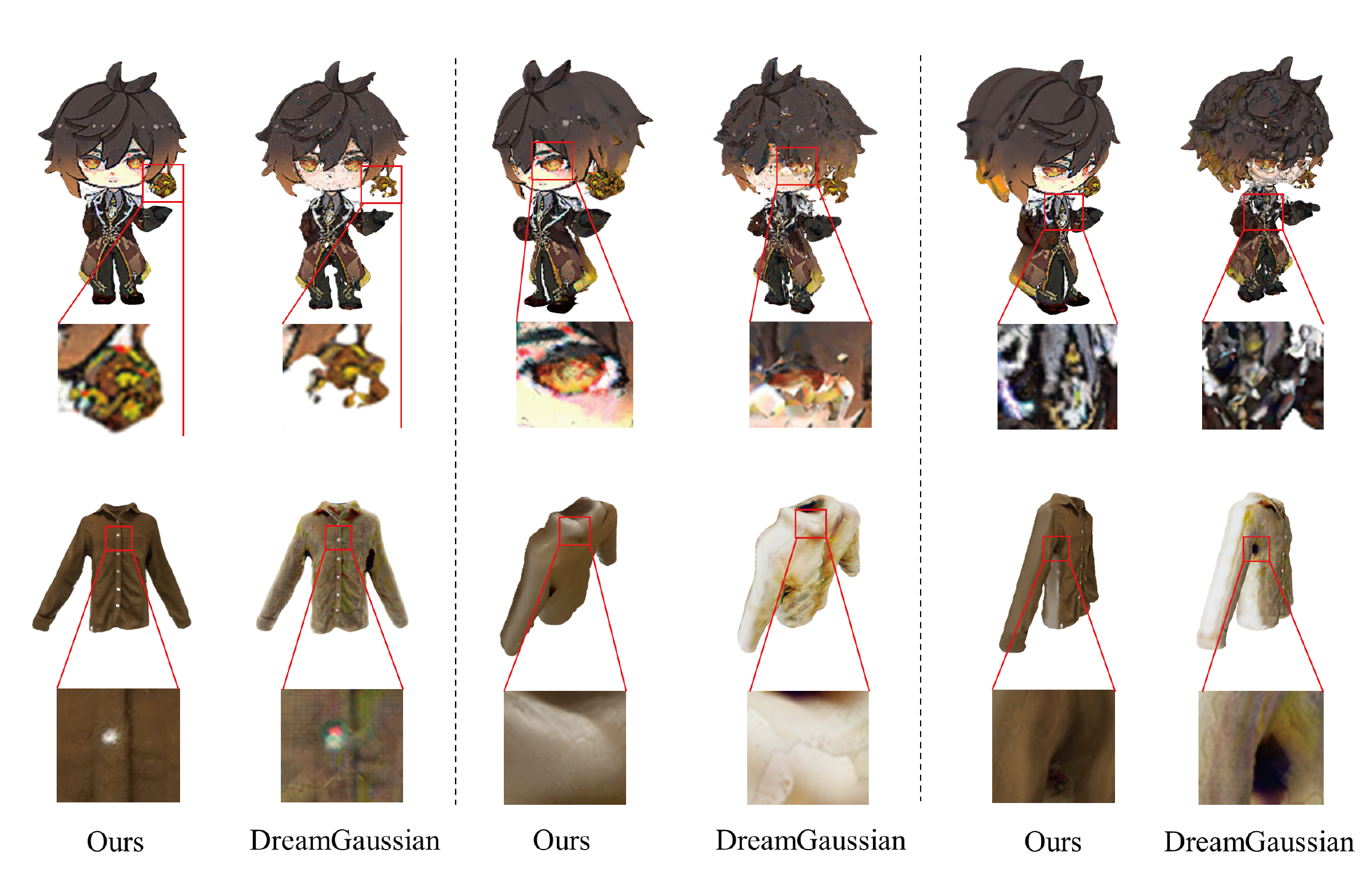}
   \caption{We propose a 3D Content Creation framework, called ScalingGaussian, which efficiently optimizes 3D Gaussian Splatting through 2D and 3D diffusion produces 3D assets with rich details and consistent structure.}
   \label{fig:mov}
 \end{figure*}
\section{Introduction}\label{intro}

The generation of 3D assets encompasses a multitude of applications, including the preservation of digital heritage, the entertainment industry, and the field of robotics. The creation of 3D assets typically necessitates the expertise of highly skilled professionals who employ specialized software to achieve the desired level of quality. This process involves the application of sophisticated techniques and tools that enable the precise modeling, texturing, and rendering of 3D objects. The expansion of the gaming and virtual reality (VR) sectors, along with the widespread adoption of 3D modeling tools, has led to a corresponding increase in the demand for 3D assets.

The advent of technologies such as image-to-3D \cite{long_wonder3d_2023,tang_dreamgaussian_2023} and text-to-3D \cite{chen_text--3d_2024,poole_dreamfusion_2022,yi_gaussiandreamer_2023} has contributed to a reduction in the cost of professional input, thereby enabling the creation of 3D content by non-professionals. The primary objective of image-to-3D and text-to-3D development is the rapid generation of highly detailed 3D content, which holds significant potential for democratizing the production of 3D assets and broadening their accessibility across various domains.


The generation of 3D content can be approached through two principal methodologies: 3D native methods and 2D-based optimization methods. In 3D native methods, models such as Shap-E \cite{jun2023shap} and Point-E \cite{nichol2022point} train a 3D diffusion model utilizing 3D data. These 3D diffusion models exhibit robust generalization capabilities, enabling the generation of highly consistent 3D assets from a single image. However, the scarcity of 3D training data sources has resulted in the formation of 3D datasets that are typically smaller and less diverse than their 2D counterparts. This limitation reduces the capacity of Shap-E and Point-E to generate intricate texture details.

Projection-Conditioned Point Cloud ($\text{PC}^2$) Diffusion \cite{melas-kyriazi_pc2_2023} employs a single image and a camera pose as conditions to constrain the diffusion process of point clouds. At each diffusion time step, it adjusts the process by projecting the image onto the denoised point cloud, thereby achieving greater geometric consistency and some degree of texture detail. Nevertheless, significant challenges persist due to the elevated computational costs associated with 3D data and the intrinsic complexity of handling 3D data. These obstacles highlight the need for continued innovation and optimization within the field to fully harness the potential of 3D native methodologies for content generation.

The simplicity of acquiring 2D data and the abundance of diverse 2D training datasets enable 2D diffusion models a distinct benefit in generating complex content\cite{saharia2022palette,rombach2022high}. Taking advantage of the abundant data available in 2D diffusion, \cite{tang_dreamgaussian_2023,poole_dreamfusion_2022} can generate detailed textures and complex geometric shapes based on input images and poses. 
Although 2D-based optimization methods leverage 2D diffusion to address the issue of missing details, the inherent randomness of 2D diffusion results in the generation of 3D assets that exhibit poor geometric consistency.

In order to achieve optimal geometric consistency in the generated three-dimensional assets while preserving texture details, this paper utilizes a 3D Gaussian-based approach that combines the strengths of both 3D diffusion and 2D diffusion.
The initial stage of the process entails the utilization of 3D diffusion\cite{nichol2022point}, which is employed to generate a 3D asset, the resulting data of which is outputted as a point cloud. 
Subsequently, this initial point cloud serves as the input for the 3D Gaussians. 
However, the point cloud generated by 3D diffusion for a single object tends to be overly sparse, resulting in an insufficient number of initialized 3D Gaussians. 
This negatively impacts the structure of the final 3D asset.

In the paper, we further propose a method to densify the initialization of 3D Gaussians. 
The method is composed of two principal modules: a scaling module and a perturbation module. 
The scaling module involves selecting local regions, generating points according to a uniform distribution, and employing local density-weighted selection to retain points.
In the event that no initialized Gaussians are available within a specified area, the process of optimizing new Gaussians points within this region during subsequent iterations becomes a significant challenge. 
This can reduce the overall efficiency of the generation framework.
To ensure that each region in space has an initial point, a perturbation module is also incorporated with the objective of improving the efficiency of the generation process.
The perturbation module uniformly generates random points in space and estimates the densest color to assign to these random points.
Since the input consists of a single-view image, we utilize 2D diffusion to assist in optimizing the 3D Gaussians. 
We leverage the Score Distillation Sampling (SDS) \cite{poole_dreamfusion_2022} loss to interact with the 2D diffusion model, guiding the 3D Gaussians to further clone and split.
To enhance the texture details of the 3D content, we extract the mesh from the final 3D Gaussians and optimize the textures using Mean Square Error(MSE) and Gradient Profile Prior(GPP) \cite{sun2010gradient}.

The contributions of our work can be summarized as follows:
\begin{itemize}
\item To achieve good geometric consistency and detailed textures in 3D assets, we propose a novel 3D content creation framework that integrates the strengths of both 3D diffusion and 2D diffusion. We interact with the 2D diffusion model using SDS loss. It can guide the 3D Gaussians to further clone and refine, thereby enhancing the overall quality of the 3D assets.
\item We propose a densification method to address the issue of sparse point clouds generated by 3D diffusion for individual objects. The method consists of a scaling module and a perturbation module. The scaling module enhances the geometric structure features generated by the model, while the perturbation module improves generation efficiency.
\item Experiments on image-to-3D tasks have shown that our method is efficient and effective. It can generate a 3D asset in under 90 seconds on a single GPU. This method demonstrates significant potential in achieving geometric consistency and detailed textures in 3D assets generation.
\end{itemize}






\section{Related Work}

\subsection{3D Representations}
Representations of 3D assets can be categorized into implicit representations and explicit representations.
The former defines 3D shapes through implicit functions without directly storing 3D geometric information. Implicit representations are typically more compact and capable of representing complex topological structures\cite{sundermeyer_learning_nodate,barron_mip-nerf_2021,zhang_nerf_2020,martin-brualla_nerf_2021,mildenhall_nerf_nodate}.
Neural Radiance Fields (NeRF)\cite{mildenhall_nerf_nodate} are a typical method of implicit representation, enabling the generation of high-quality 3D content with known camera parameters. NeRF employs a multi-layer perceptron (MLP) to map 3D coordinates and viewing directions to color and density, and generates new views through volume rendering.
NeRF has limitations in handling high-frequency details, often leading to aliasing artifacts.
Mip-NeRF (Multiscale Representation) \cite{barron_mip-nerf_2021} addresses this issue by introducing a multiscale feature, i.e.,  
replacing the rendering rays with anti-aliased conical frustums to effectively mitigate the blurring problem when dealing with high-frequency details.
DSNeRF\cite{kangle2021dsnerf} incorporates depth supervision alongside color supervision, allowing scene reconstruction with fewer input views and faster training speeds compared to NeRF.
Although NeRF-based methods can generate high-quality and highly realistic 3D content, they are computationally intensive, resulting in very time-consuming training and rendering processes\cite{poole_dreamfusion_2022,barron_mip-nerf_2021,mildenhall_nerf_nodate}.

Recently, 3D Gaussian Splatting\cite{kerbl_3d_2023} has emerged as the mainstream 3D representation method, replacing NeRF. As an explicit representation, 3D Gaussian Splatting can achieve results comparable to NeRF. It has demonstrated impressive speed in image-to-3D tasks and does not require the extensive training time that NeRF does. Furthermore, 3D Gaussian Splatting provides a more intuitive representation of geometric shapes, making the generated 3D content easier to edit.
SparseGS\cite{xiong_sparsegs_2023} enhances 3D Gaussian Splatting by incorporating depth supervision from rendered images. This approach enables scene reconstruction with fewer images and improves rendering accuracy and efficiency.
Due to the superior performance of 3D Gaussian Splatting, our method also employs 3D Gaussians as the representation technique, which  enables us to generate detailed 3D assets from single-view images in a significantly short amount of time.

\subsection{Image to 3D generation}
The field of image-to-3D generation has made remarkable strides in recent years, driven by advancements in deep learning and computer vision technologies. The task of Image-to-3D generation involves producing 3D assets from 2D images\cite{abdal_gaussian_2023,xiong_sparsegs_2023,lu_scaffold-gs_2023}. 
Early image-to-3D generation primarily relied on Structure from Motion (SfM) and Multi-View Stereo (MVS) techniques. 
SfM \cite{schonberger2016structure, wang2021deep} estimates the 3D structure of a scene by analyzing the motion between several images taken from different viewpoints and generates a sparse 3D point cloud representing the scene.
The most classic approach in MVS is MVSNet\cite{yao2018mvsnetdepthinferenceunstructured}, which  introduces a learning-based method that utilizes convolutional networks to extract deep image features from multi-view images. 
LE-MVSNet \cite{10.1007/978-3-031-44198-1_40} addresses the high computational costs of MVSNet by integrating depth feature extraction and edge feature extraction modules. This enhances resource utilization efficiency without compromising model performance.
The above methods require multiple images taken from different angles to reconstruct a 3D scene. 
Although they achieve excellent representation results, they are computationally intensive and involve high costs for image acquisition during model training.

Directly applying SfM or MVS to a single-view image for 3D reconstruction results in highly blurred 3D content.
The primary reason for this blurriness is the lack of sufficient perspective information needed to accurately infer the 3D shape and details of an object. 
\cite{choy20163dr2n2unifiedapproachsingle} indicates that without multiple images, MVS and SfM cannot leverage the necessary parallax and feature matching, leading to poor output quality.
Consequently, the generated models struggle in producing fine details and complex structures, failing to achieve the precision that multi-view reconstruction can provide.
Zero-1-to-3 \cite{liu_zero-1--3_nodate} leverages CLIP\cite{radford_learning_2021} to apply camera transformation in 2D diffusion, enabling zero-shot image-conditioned novel view synthesis. 
SyncDreamer\cite{liu_syncdreamer_2023} uses 3D convolutions to extract spatial geometric features, which guide the denoising function of Zero-1-to-3. This approach enhances geometric consistency while preserving texture details.
DreamFusion\cite{poole_dreamfusion_2022} eliminates the need for 3D data and multi-view training data. This method connects the 2D diffusion model with the 3D representation model by introducing the score distillation sampling (SDS), which
leverages the pre-trained 2D diffusion model (Zero-1-to-3), to guide the 3D representation model duting the updating of its parameters.
By utilizing the combination of pre-trained 2D diffusion models and SDS, in this paper, our proposed method can generate detailed 3D content from a single-view image.

\section{Methods}
\begin{figure*}[ht]
   \centering
   \includegraphics[width=\linewidth]{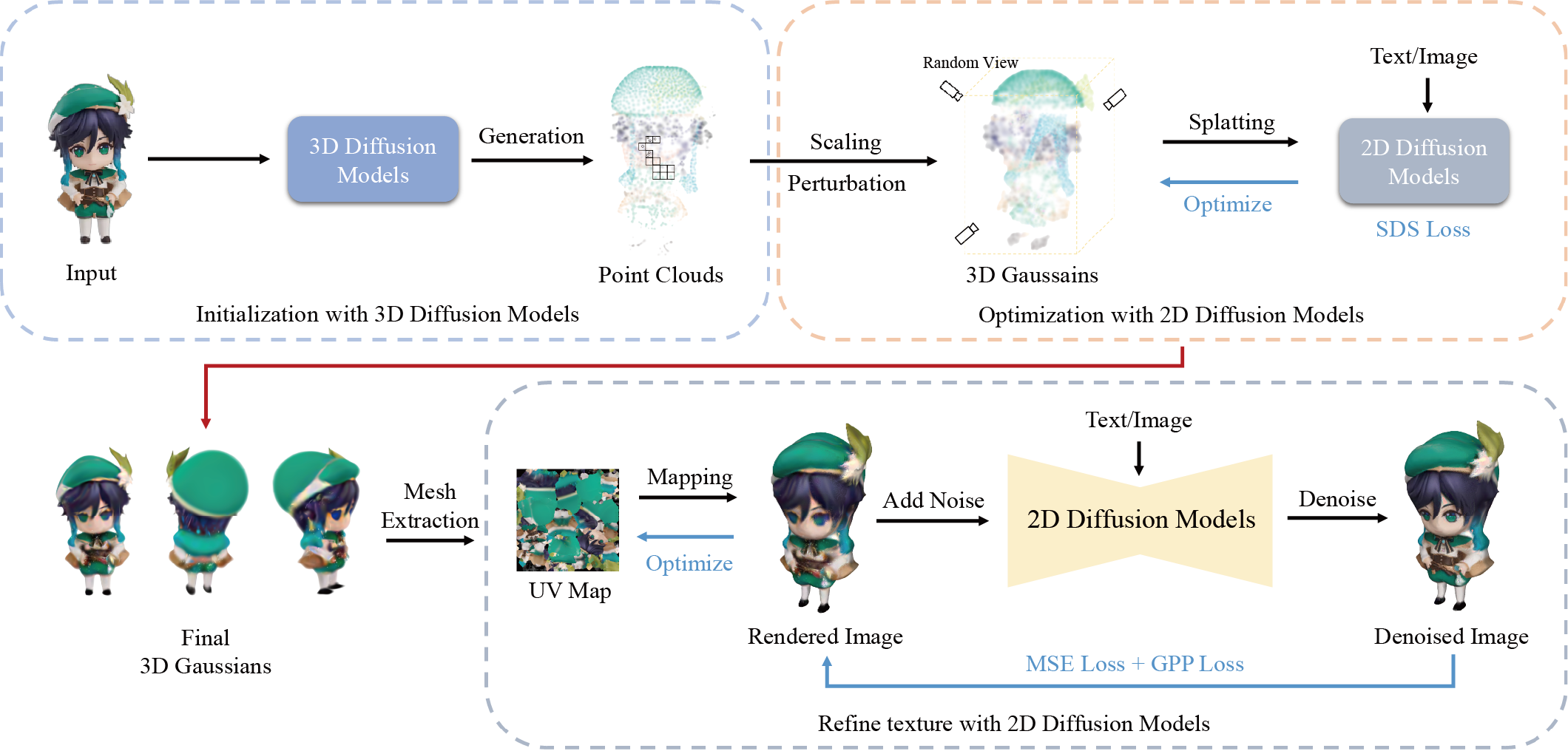}
   \caption{Overall framework. First, the 3D diffusion model generates the initial point clouds. Then, scaling and perturbation modules are applied to these point clouds to enhance the structural characteristics of the object. Scaling points are used to create initialized 3D Gaussians, which  are then optimized using SDS loss in conjunction with the 2D diffusion model. Finally, images are rendered through 3D Gaussian Splatting and a textured mesh is extracted and refined using the 2D diffusion model, where MSE loss and GPP loss are applied to enhance texture details.}
   \label{fig:Pipeline}
 \end{figure*}



In this section, we present the framework for 3D content generation employed in image-to-3D tasks, as depicted in Fig.\ref{fig:Pipeline}. 
We utilize 3D Gaussian Splatting as the representation method, optimizing their parameters through the use of the SDS loss. A detailed review of 3D Gaussian Splatting will be provided in Section \ref{sub3d}.
The sparsity of point clouds produced by 3D diffusion can diminish the overall efficiency of the generation framework. Consequently, densifying the initial 3D Gaussians becomes essential. In Section \ref{scalper}, we will discuss how the scaling and perturbation modules are employed to enhance these initial 3D Gaussians.
The optimization of 3D Gaussian-rendered images solely through the use of SDS inevitably results in the blurring of image edges. 
In Section \ref{reftexture}, we enhance texture features by converting 3D Gaussians into meshes and then applying the GPP loss, which results in textures with significantly sharper edges.
\subsection{Formulation}
In the context of a 3D reconstruction task, the input is typically constituted by a set of images and point clouds, which are generated through the application of the SfM or 3D diffusion. \( P = \{ p_1, p_2, \ldots, p_N \} \) represents the sparse point clouds containing \( N \) points, and \( I = \{ I_r, I_{\alpha}\} \) denotes the set of \( 2 \) images. Each point \( p_i \) in the point clouds \( P \) corresponds a specific coordinate in the 3D space. The quality of the 3D reconstruction is dependent on the density and distribution\( P \) of the point clouds. A sparse point cloud can result in an inadequate geometric representation, which negatively affects the accuracy and detail of the reconstructed model.
\subsection{Preliminaries}\label{sub3d}
\subsubsection{3D Gaussian Splatting}
The method of 3D Gaussian Splatting employs anisotropic Gaussians to represent the scene. 
These 3D Gaussians are initially derived from SfM point clouds.
Each gaussian$ \{ \mathcal{G}_k \mid k = 1, \ldots, K \}$
is characterized by its center position $\mu_k \in \mathbb{R}^{3\times1}$, covariance $\Sigma_k \in \mathbb{R}^{3\times3}$, color $c_k \in \mathbb{R}^{3}$, and opacity $\alpha_k \in \mathbb{R}^1$, rotation factor $r_k \in \mathbb{R}^{3}$, and scaling factor $q_k \in \mathbb{R}^{4}$:
\begin{equation}
 \mathcal{G}_k(\delta x) = e^{-\frac{1}{2}(\delta x)^T \Sigma^{-1}_k (\delta x)},
\end{equation}
where $\delta x = x-\mu_k$, $x$ represents an arbitrary position in the 3D space and $\delta x$ is the distance from the center position $\mu_k$. The covariance matrix $ \Sigma_k $ is calculated from the rotation matrix $ R_k $ and the scaling matrix $ S_k $. The covariance matrix $ \Sigma_k $ must be positive semi-definite to enable differentiable optimization:
\begin{equation}
\Sigma_k = R_k S_k (R_k S_k)^T.
\end{equation}

When rendering new viewpoints, the 3D Gaussians need to be projected into 2D. 
This requires the use of a viewing transformation $W_k$ and the Jacobian matrix $J_k$ of the approximate affine transformation. The 3D Gaussians are first transformed into camera coordinates and then projected into ray space:
\begin{equation}
\mu'_k = W_k \mu_k,\Sigma'_k = J_k W_k \Sigma (J_k W_k)^T,
\end{equation}
where the Jacobi matrix $J_k$ is defined by the 3D Gaussian center $\mu'_k$.

To calculate the color of each pixel, 3D Gaussian Splatting employs neural point-based rendering. First, the camera center is determined, and a ray $r$ is cast from the camera center. 
The 3D Gaussian $\mathcal{G}_k$ is transformed into a 2D Gaussian $\mathcal{G}'_k$ in the image plane\cite{zwicker2001ewa}.
Then, the color and density of the 3D Gaussians intersected by the ray $r$ are computed. The formula for sampling $ N $ points intersected by the ray $r$ is as follows:
\begin{equation}
C_r(\delta x) = \sum_{i \in N} c_i \sigma_i \prod_{j=1}^{i-1} (1 - \sigma_j), \quad \sigma_i = \alpha_i \mathcal{G}'_i(\delta x),
\end{equation}
where $ c_i $ and $ \alpha_i $ represent the color and opacity of the $i$-th Gaussian, respectively. $ c_i $ is computed using the spherical harmonics (SH) function.

By reconstructing images from training viewpoints, all attributes of the 3D Gaussians can be optimized. The loss function is defined as follows:

\begin{equation}
\mathcal{L} = (1 - \lambda) \mathcal{L}_{1} + \lambda\mathcal{L}_{\text{D-SSIM}}.
\end{equation}

Finally, an adaptive Gaussian densification scheme is used to clone and prune the 3D Gaussians.

\subsection{Gaussian Initialization}\label{scalper}
\subsubsection{Scaling module}
\begin{figure}
    \centering
    \includegraphics[width=\linewidth]{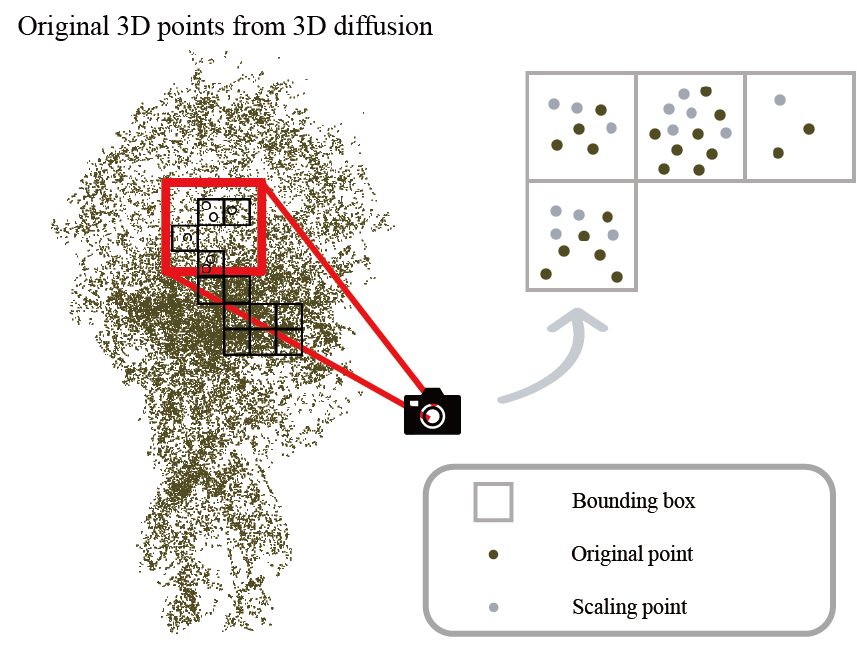}
    \caption{Scaling points based on the intrinsic characteristics of the object structure. The point cloud is divided into multiple local regions, with each region processed separately to enhance point density and color retention. Initially, new points are uniformly generated within each region, and Gaussian noise is added to the colors of the nearest neighbor points. Subsequently, a density function is constructed to retain points based on their local density.}
    \label{fig:scaling module}
\end{figure}

Given point clouds $ p_m $ by 3D diffusion, we divide it into several local regions $N$. The bounding box of each region is defined as:
\begin{equation}
     p_{\min_i}=\min(p_{m_i}),\quad p_{\max_i} = \max(p_{m_i}),\quad \forall i \in \{1, \ldots, N\},
\end{equation}
where  $p_{m_i}$  represents the points in the  $i$ -th region, and $ N $ is the number of regions.
The position of the newly generated point $ p_{r_i} $ follows a uniform distribution within the bounding box:
\begin{equation}
       p_{r_i} \sim \text{Uniform}(p_{\min_i}, p_{\max_i}).
\end{equation}
The color of the newly generated point $ c_{r_i} $ is obtained by adding Gaussian noise to the color of the nearest neighbor point $ c_{m_i} $:
\begin{equation}
     c_{r_i} =  c_{m_i} + \gamma, \quad \gamma \sim \mathcal{N}(0, \sigma_s^2).
\end{equation}

Now, we can then construct the density function $ \rho(p) $ of the point cloud, and select the points to be retained based on the density-weighted method:
\begin{equation}
    \rho(p) = \sum_{i=1}^k \exp\left(-\frac{\|p - p_i\|^2}{2\sigma_p^2}\right),
\end{equation}
where $ p_i $ are the $ i $ nearest points in the point cloud.  
The calculation method involves finding the \( k \) nearest neighbors \( p_i \) of point \( p \). 
In each region, we construct a KDTree\cite{bentley1975multidimensional} to search for points.
The distances between the points and \( p \) are computed and then converted into density values using a Gaussian function. 
The probability of selecting a retained point is proportional to the local density:
\begin{equation}
    \mathbb{P}(p_{r_j} \in p_{r_{i}}) \propto \rho(p_{r_j}).
\end{equation}
The probability \( \mathbb{P}\) of a point \( p_{r_j} \) being selected as an effective point is proportional to the density \( \rho(p_{r_j}) \). The higher the density, the greater the probability of the point being selected(see Fig.\ref{fig:scaling module}).

\subsubsection{Perturbation Module}
To ensure that each initialized region contains a sufficient number of 3D Gaussians, we introduce a perturbation module. We generate \(M\) random points \(\mathbf{p}_{\text{noise}}\) within a sphere of radius \(r\) of the initialized point. Using spherical coordinates \((\phi, \beta , \psi)\), these points are calculated as:

\begin{equation}
\mathbf{p}_{noise} :=
\begin{cases}
p_x \\
p_y \\
p_z 
\end{cases}
=
\begin{cases}
r \psi^{1/3} \sin \beta \cos \omega \\
r \psi^{1/3} \sin \beta  \sin \omega \\
r \psi^{1/3} \cos \beta
\end{cases},
\end{equation}
where \(\omega \in [0, 2\pi)\), \(\beta  \in [0, \pi]\), and \(\psi \in [0, 1)\) are uniformly distributed random variables. Each generated noise point is assigned the most densely populated (densest)  color \( c_{dense} \).
To determine the densest color in the color distribution, we use Kernel Density Estimation (KDE)\cite{doi:10.1137/1114019}, which   is employed to estimate the probability density function of a random variable. Given color data \( C = \{ \mathbf{c}_1, \mathbf{c}_2, \ldots, \mathbf{c}_n \} \), 
KDE estimates the probability density function \(\hat{f}(\mathbf{c})\) of the color distribution:
\begin{equation}
   \hat{f}(\mathbf{c}_i) = \frac{1}{n h^d} \sum_{j=1}^{n} K \left( \frac{\mathbf{c}_i - \mathbf{c}_j}{h^d} \right),\quad \forall i \in \{1, \ldots, n\},
\end{equation}
where \(K\) is the Gaussian kernel, \(h^d\) is the bandwidth, and \(\mathbf{c}_j\) are the color samples. By calculating the log density of each color sample point, we identify the sample point with the highest density value. The corresponding color is then considered to be the densest color:
\begin{equation}
     \mathbf{c}_{\text{dense}} = \arg \max_{\mathbf{c}_i} \log \hat{f}(\mathbf{c}_i).
\end{equation}

Each generated noise point is assigned the densest color \(\mathbf{c}_{\text{dense}}\):
\begin{equation}
    \mathbf{c}_{\text{noise}} = \{ \mathbf{c}_{\text{dense}} \}_{i=1}^M.
\end{equation}

\subsection{Density in Optimization}\label{optim}

The role of parameter initialization in 3D Gaussian Splatting is pivotal for effective optimization. By defining the optimization parameters \(\theta = (\mu, c, s, q, \alpha)\), the goal is to minimize a loss function that aligns reference view images and transparency with input images. The optimization employs gradient descent, iterating to update \(\theta\) and leveraging the smoothness of the loss function to ensure convergence. Crucially, the choice of the initial point \(\theta_0\) impacts the convergence speed and the likelihood of finding a global optimum. Proximate initial points to the optimal solution can accelerate convergence and reduce iterations, while inappropriate initial points may lead to suboptimal local minima. Therefore, initializing parameters with a close estimate to the optimal solution is essential for efficient and effective optimization.


\subsubsection{The Role of Parameter Initialization}
In 3D Gaussian Splatting, the optimization parameters include \(\mu\), \(c\), \(s\), \(q\), and \(\alpha\). These parameters can be combined into a single vector \(\theta\), which is expressed as:
\begin{equation}
    \theta = (\mu, c, s, q, \alpha).
\end{equation}
The optimization objective is to minimize a specific loss function that aligns the reference view image \(I^{r}_{R}\) and transparency \(I^{r}_{\alpha}\) with those of the input images \(\tilde{I}^{r}_{R}\) and \(\tilde{I}^{r}_{\alpha}\), respectively. 
Additionally, other object viewpoints optimized for SDS loss have been considered.
The loss function $\mathcal{L}$ is defined as follows:
\begin{equation}
\begin{aligned}
    \min_{\theta} &\mathcal{L}_r(\theta) 
    = \min_{\theta}  \lambda_{R} \underbrace{ \| I^{r}_{R}(\theta) - \tilde{I}^{r}_{R} \|_2^2 }_{\text{RGB loss of know view}} \\
    &+ \lambda_{\alpha} \underbrace{ \| I^{r}_{\alpha}(\theta) - \tilde{I}^{r}_{\alpha} \|_2^2}_{\text{Foreground mask loss of know view}} \\
    &+ \lambda_{s} \underbrace{ \mathbb{E}_{t,\Delta r,\epsilon} \| \epsilon_{\phi}(I_{R}^{r+\Delta r}(\theta); t, \tilde{I}_{R}^{r},\Delta r) - \epsilon \|_2^2 }_{\text{SDS loss of novel view}}, \\
\end{aligned}
\end{equation}
where $\epsilon_{\phi}(.)$ represents the predicted noise by the 2D diffusion prior \(\phi\), \(\Delta r\) denotes the relative camera pose change from the reference camera \(r\), and $t$ represents the time step in 2D diffusion.
To accomplish this, we utilize the gradient descent method\cite{kingma2014adam}, iterating according to the following formula:
\begin{equation}\label{grad}
    \theta_{k+1} = \theta_k - \eta \nabla \mathcal{L}_r(\theta_k),
\end{equation}
where \(\eta\) represents the learning rate and \(\nabla \mathcal{L}_r(\theta_k)\) denotes the gradient of \(\mathcal{L}_r\) evaluated at \(\theta_k\). Assuming the loss function \(\mathcal{L}_r(\theta)\) is \(L\)-smooth(The proof of the smoothness of the loss function can be found in \ref{appen}), we have for any \(\theta\) and \(\theta'\):
\begin{equation}
    \mathcal{L}_r(\theta') \leq \mathcal{L}_r(\theta) + \nabla \mathcal{L}_r(\theta) \cdot (\theta' - \theta) + \frac{L}{2} \|\theta' - \theta\|^2. 
\end{equation}
By substituting the gradient descent iteration Eq.\ref{grad} into this inequality, we obtain:
\begin{equation}
    \mathcal{L}_r(\theta_{k+1}) \leq \mathcal{L}_r(\theta_k) - \left( \eta - \frac{L \eta^2}{2} \right) \|\nabla \mathcal{L}_r(\theta_k)\|^2. 
\end{equation}
To ensure that the loss function \(\mathcal{L}(\theta)\) decreases at each iteration, the following condition must be satisfied:
\begin{equation}
    a = \eta - \frac{L \eta^2}{2} > 0 ,
\end{equation}
\begin{equation}\label{final}
    \mathcal{L}_r(\theta_{k+1}) \leq \mathcal{L}_r(\theta_k) - a\|\nabla \mathcal{L}_r(\theta_k)\|^2.
\end{equation}

According to Eq.\ref{final}, the initial point $\theta_0$ plays a crucial role in the optimization process. The proximity of the initial point to the optimal solution affects the convergence speed, i.e.,  closer initial points lead to faster convergence. The selection of an optimal initial point can reduce the number of iterations required. Secondly, different initial points may lead the algorithm to converge to different local minima. Thus, the choice of an appropriate initial point increases the likelihood of finding the global optimum. Additionally, the gradient magnitude at the initial point influences the initial convergence speed, where a larger gradient can accelerate the initial convergence.

\subsubsection{Structural Similarity}
The prior point cloud incorporates the geometric information of an object surface, providing an initial estimate that closely matches the actual object geometry. In contrast, random initialization lacks such geometric information, and potentially causes the initial geometric structure to deviate significantly from the true object surface. This increases the risk of becoming trapped in local optima during optimization\cite{9882220}. 

To quantify the impact of initialization quality on the optimization process, we define a loss function based on the average distance between the point cloud and the target geometric structure. 
Given two sets of points \( P = \{p_1, p_2, \ldots, p_m\} \) and \( Q = \{q_1, q_2, \ldots, q_n\} \), where \( P \) represents the predicted point positions of the model and \( Q \) represents the actual points scanned, the mean distance \( D_{\text{mean}} \) can be expressed as follows:
\begin{equation}
    D_{\text{mean}} = \frac{1}{2} \left( \frac{1}{m} \sum_{i=1}^{m} \min_{j} \| p_i - q_j \| + \frac{1}{n} \sum_{j=1}^{n} \min_{i} \| q_j - p_i \| \right).
\end{equation}
In addition, the influence of the density of the point cloud on the optimization process is considered.
\cite{dai2020neural} highlights the crucial role of density in achieving optimal final rendering effects.
Dense point clouds can describe the surface of object more accurately, offering better geometric information when using Gaussian distribution for initialization.
Dense point cloud prior information helps preserve geometric consistency during the optimization process, minimizing the effects of information loss and error propagation associated with sparse point clouds.

\subsection{Refine texture}\label{reftexture}
\begin{figure}
    \centering
    \includegraphics[width=\linewidth]{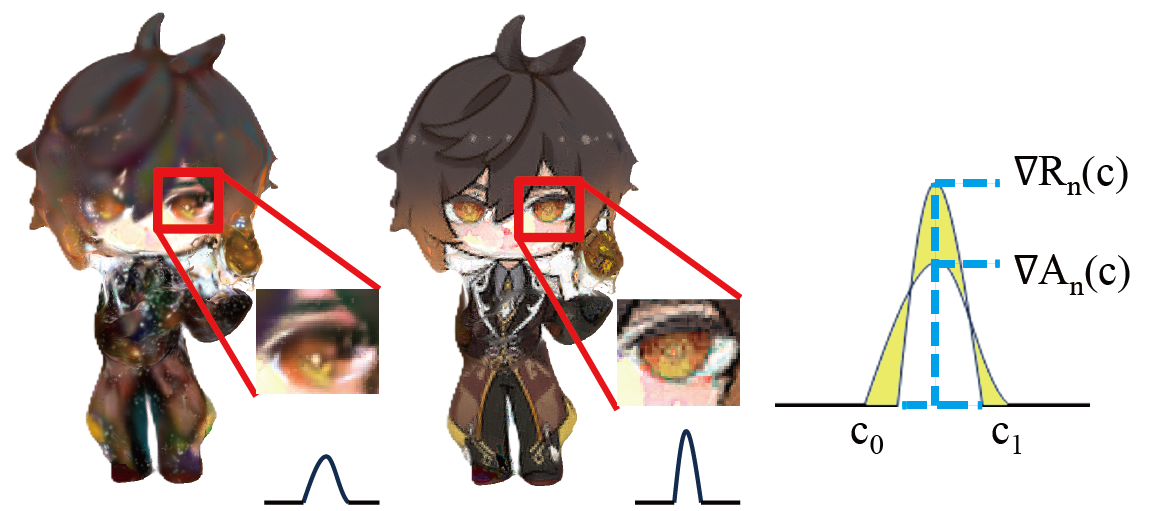}
    \caption{Gradient field and Gradient Profile Prior Loss. The gradient field of a clear image is sharper than that of a blurred image.}
    \label{fig:Refine texture}
\end{figure}
Under the guidance of SDS, the optimized final 3D Gaussians form discrete geometric structures. Unlike continuous representations of NeRF, these discrete structures are not as easily converted into polygonal meshes.
We employ an algorithm that utilizes block-wise local density query and color back-projection to extract the meshes\cite{tang_dreamgaussian_2023}.
The extracted meshes lack texture details and the rendered images have blurred edges\cite{liao2023tadatextanimatabledigital}.
Utilizing the capability of 2D diffusion, we  can remove noise and synthesize detailed 2D images by applying it to correct the colors of the meshes and refine the rendered images from various viewpoints.

For the reconstructed 3D content, the images $A_n$ are rendered from $N$ randomly selected viewpoints.
Add random noise $\varepsilon$ to the rendered images $A_n$ and then use 2D diffusion for denoising to obtain refined 2D images ${R}_{n}$:
\begin{equation}
    {{R}_{n}}=\Phi_{\theta} ({{A}_{n}},\varepsilon ,t),\quad \forall n \in \{1, \ldots, N\},
\end{equation}
where \(\Phi_{\theta}\) is the denoising function in 2D diffusion, and \(t\) represents the denoising time step.

To optimize the texture of the 3D content, pixel-level MSE loss($\mathcal{L}_{MSE} = \left\| {{R}_{n}}-{{A}_{n}} \right\|_{2}^{2}$) is used to enhance overall detail, and Gradient Profile Prior (GPP) is further employed to sharpen the edges, as detailed in Fig.\ref{fig:Refine texture}. 
The GPP loss is defined as follows:
\begin{equation}
    \mathcal{L}_{GPP} = \frac{1}{c_1 - c_0} \int_{c_0}^{c_1} \left\| \nabla R_n(c) - \nabla A_n(c) \right\|_1 \, dc, c \in [c_0,c_1],
\end{equation}
where  \(\nabla R_n\) represents the gradient field of \(R_n\), and \(\nabla A_n\) represents the gradient field of \(A_n\).

Our final defined loss is as follows:
\begin{equation}
    \mathcal{L}_{refine}=\mathcal{L}_{MSE}+\zeta \mathcal{L}_{GPP},
\end{equation}
where \(\zeta\) is a hyperparameter that balances the importance between MSE and GPP.


\section{Experiments}
In this section, we first introduce the dataset in \ref{dataset} and then describe the implementation details in \ref{imple}. In \ref{quality}, we present the visual results of our method and perform both quantitative and qualitative analyses compared to other methods. Finally, in \ref{Ablation}, we perform a series of ablation experiments to verify the effectiveness of each module in our method.

\subsection{Datasets}\label{dataset}
Our approach involves testing model performance using two different datasets and some photos taken on the spot with a mobile phone.
\begin{enumerate}[a)]
\item Deep Fashion3D \cite{zhu2020deepfashion3ddatasetbenchmark} comprises 2078 3D assets reconstructed from real garments, comprising 10 different categories and 563 clothing instances. Deep Fashion3D provides extensive annotations, including 3D feature lines, 3D body poses, and multi-view real images. The objective is to generate 3D models from a single-view image. From each of these categories, we randomly choose 3 instances for our experiments. For each instance, a single-view image is rendered to generate the 3D asset.
\item Animal Set \cite{10.1145/1553374.1553469} includes a collection of 60 different types of toy animals, such as horses, ducks, cows, sheep, deer, dogs, cats, pigs, mice, rabbits and various kinds of birds. Five different categories are selected from this dataset, and a single-view image from each category is chosen for our experiments.
\end{enumerate}

\subsection{Implementation Details}\label{imple}
Our method is implemented using the PyTorch framework. For the entire framework, testings are conducted after 600 iterations. For 3D diffusion, the pre-trained Point-e model generates point clouds of objects. Each point cloud produced by Point-E consists of 4096 points. For 2D diffusion, we utilize Zero-1-to-3 with a guidance scale of 5. 
For 3D Gaussians,the position $\mu$ for training begins with an initial learning rate of 0.001, which gradually decreases to a final learning rate of 0.00002. 
The covariance of the 3D Gaussian is derived from scaling and rotation, so only scaling and rotation need to be optimized, rather than directly optimizing the covariance. 
The 3D Gaussian training parameters are opacity learning rate 0.05, scaling learning rate 0.005, and rotation learning rate 0.005. 
The color $c$ of the 3D Gaussian is represented using spherical harmonic (SH) coefficients. The SH coefficients are set to gradually increase from an initial value of 0 to 3, with a learning rate of 0.01.
When rendering new viewpoints, the camera pose is randomly sampled with a fixed radius of 2, an azimuth angle ranging from -180 to 180 degrees, and an elevation angle ranging from -60 to 60 degrees.
During the 600 iterations, the density of the 3D Gaussian is adjusted at intervals of 50 steps.
All experiments are conducted and measured on an RTX 4090 (24GB) GPU.

\subsection{Quantitative comparisons}\label{quality}

\begin{table}[t]
\caption{Quantitative comparisons on CLIP\cite{radford_learning_2021,schuhmann2022laion5bopenlargescaledataset} similarity with other methods.}
\centering
\begin{tabularx}{0.45\textwidth}{l*{3}{>{\centering\arraybackslash}X}}
\toprule
 & \textbf{ViT-B/32} & \textbf{ViT-L/14} & \textbf{ViT-bigL/14}\\
\midrule
Point-E\cite{nichol2022point} & 74.28 & 61.35 & 66.08\\
Shap-E\cite{jun2023shap} & 83.56 & 69.59 & 73.22\\
Dreamfusion\cite{stable-dreamfusion} & 88.26 & 71.78 & 74.56\\
DreamGaussian\cite{tang_dreamgaussian_2023} & 85.71 & 66.60 &72.09\\
Ours & 90.63 & 82.97 &78.22\\
\bottomrule
\end{tabularx}
\label{table:clip}
\end{table}

\begin{table*}[th]
\caption{Quantitative comparison of the generative quality for the single-image to 3D task on Deep Fashion3D.}
\centering
\begin{tabular}{lccccc}
\toprule
 & \textbf{Shap-E}\cite{jun2023shap} & \textbf{Point-E}\cite{nichol2022point} & \textbf{Dreamfusion}\cite{stable-dreamfusion} & \textbf{DreamGaussian}\cite{tang_dreamgaussian_2023} & \textbf{Ours} \\
\midrule
Hausdorff & 1.16 & 0.50 & 0.71 & 0.73  & 0.65\\
Mean Distance & 0.35 & 0.21 & 0.21 & 0.23 & 0.18\\
Generation Time & 19 seconds & 24 seconds & 1 hour & 56 seconds & 82 seconds\\
\bottomrule
\end{tabular}

\label{table:dis}
\end{table*}

\begin{figure*}[h]
   \centering
   \includegraphics[width=\linewidth]{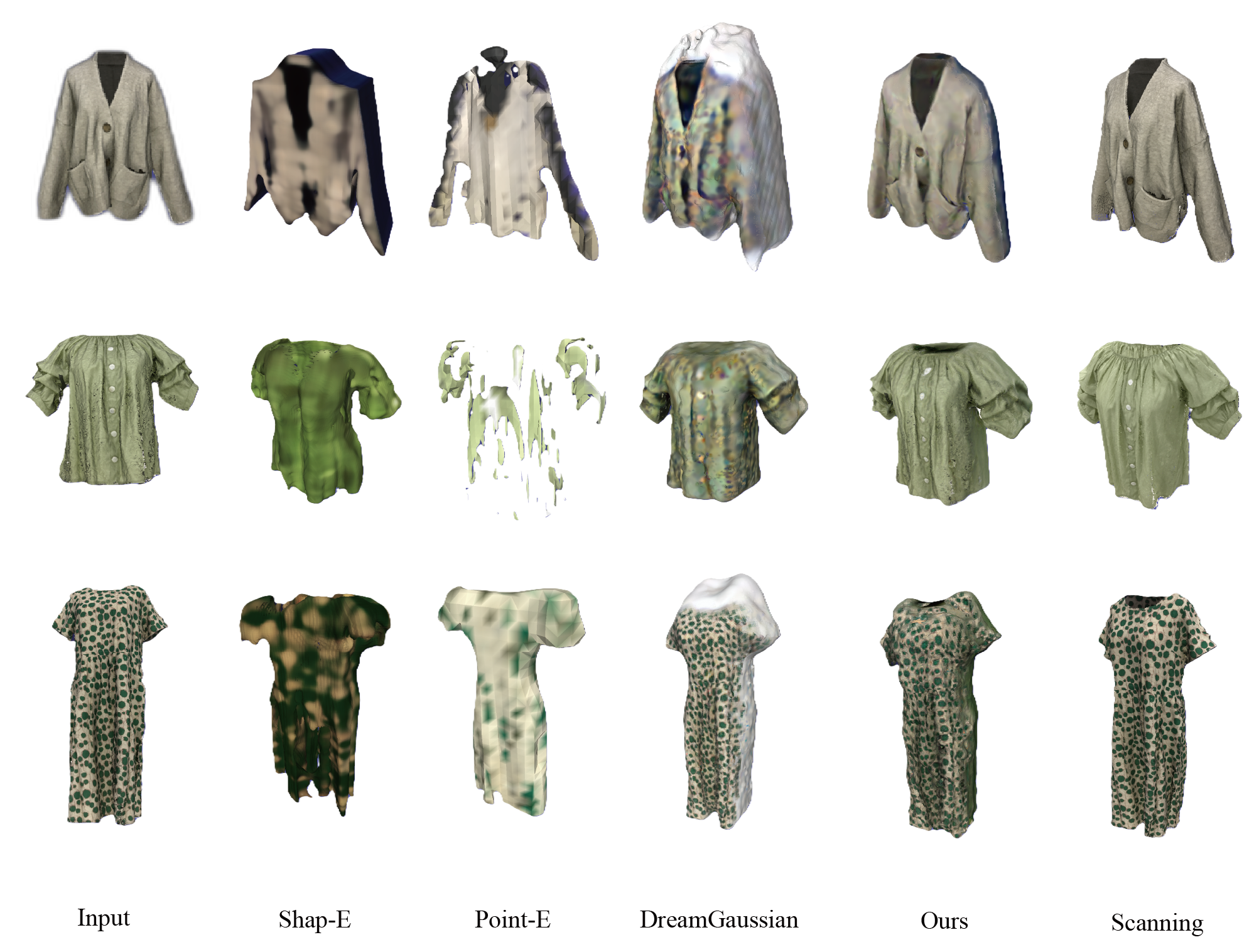}
   \caption{Qualitative comparisons of the generative quality on Deep Fashion3D.}
   \label{fig:ex}
 \end{figure*}

Due to the different methods of representing 3D content, we convert the 3D assets generated by all methods into both point cloud and mesh formats for comparative experiments.
In Tab.\ref{table:clip}, we utilize CLIP\cite{radford_learning_2021} to quantitatively assess the similarity between the newly rendered perspective images produced by each method and the input images.
During the evaluation process, we use a camera radius of 2 to render 8 images, including front, side, and top views. For each method, we calculate the similarity between each rendered image and the input image, and then compute the average similarity across all eight images.
Three models are used to evaluate similarity: ViT-B/32 and ViT-L/14 from \cite{radford_learning_2021}, and ViT-bigL/14 from OpenCLIP\cite{schuhmann2022laion5bopenlargescaledataset,ilharco_gabriel_2021_5143773}.
Based on the evaluations of the three models, our method generates different perspectives with a similarity that outperforms all other competing methods.

In Tab.\ref{table:dis}, Hausdorff distance \cite{hausdorff1914grundzuge} and Mean distance are selected as evaluation metrics. Hausdorff distance measures the maximum geometric difference, indicating the distance between the farthest point pairs. On the other hand, Mean distance provides an overall measure of geometric similarity, reflecting the average error between all point pairs. 
By using these two metrics in combination, the similarity of two 3D contents can be observed at both local and global scales.
The results indicate that our model generates 3D content with a superior geometric structure at a global scale compared to other methods. 
Similar to Dreamfusion, we utilize SDS loss in conjunction with 2D diffusion to optimize our model. However, Dreamfusion employs NeRF as its representation method, which results in slower 3D asset generation. In contrast, our model generates 3D content more efficiently.

\subsection{Qualitative Comparisons}

In Fig.\ref{fig:ex}, we present the image-to-3D results of our method compared to other competing methods on the Deep Fashion3D dataset. All competing methods are converted into polygon meshes with textured images.
In terms of generation speed, our method is over 30 seconds slower than DreamGaussian due to the incorporation of 3D diffusion. Shap-E and Point-E, which only require inference and do not involve 2D diffusion, generate results more quickly. Similar to Dreamfusion, our method also uses SDS loss in conjunction with 2D diffusion. However, Dreamfusion employs NeRF for 3D content representation, resulting in lower generation efficiency compared to our approach. 
In terms of generation quality, the 3D asset generated by Point-e displays distinct characteristics of the garment but suffers from noticeable artifacts and a lack of detail. While the overall shape is somewhat preserved, the texture and finer details are significantly lacking. 
The 3D asset produced by DreamGaussian shows some improvement in detail but still exhibits substantial artifacts and blurring. DreamGaussian uses randomly initialized Gaussian points, which make it difficult to optimize the points.
The model generated by our method demonstrates significant improvement in both geometric structure and texture quality. The overall shape of the garment is well preserved and the details are more accurate.

We also conducted a user study to evaluate the quality of our generation, as detailed in Tab.\ref{Tab:UserStudy}. This study focuses on assessing two critical aspects in the context of image-to-3D tasks: reference view consistency and overall generation quality. These metrics are essential for determining the effectiveness and accuracy of the generated 3D models.

\begin{table*}[h]
\caption{A comparative user study on image-to-3D tasks, with ratings on a scale from 1 to 5 (higher scores indicate better performance). Our method demonstrates superior performance across all evaluated criteria.}
\centering
\begin{tabular}{lccccc}
\hline
  & \textbf{DreamGuassian}\cite{tang_dreamgaussian_2023} & \textbf{Shap-E}\cite{jun2023shap} & \textbf{Point-E}\cite{nichol2022point} & \textbf{SyncDreamer}\cite{liu_syncdreamer_2023} & \textbf{Ours} \\
\hline
Ref. View Consistency $\uparrow$   & 3.35 &2.11 & 1.85& 3.19 & 4.41 \\
Overall Model Quality $\uparrow$   & 3.21 &1.92 & 1.57& 2.96 & 4.05 \\
\hline
\end{tabular}
\label{Tab:UserStudy}
\end{table*}

\subsection{Ablation Study and Analysis}\label{Ablation}
\subsubsection{Initialization with 3D Diffusion Model}
\begin{figure}[t]
   \centering
   \includegraphics[width=\linewidth]{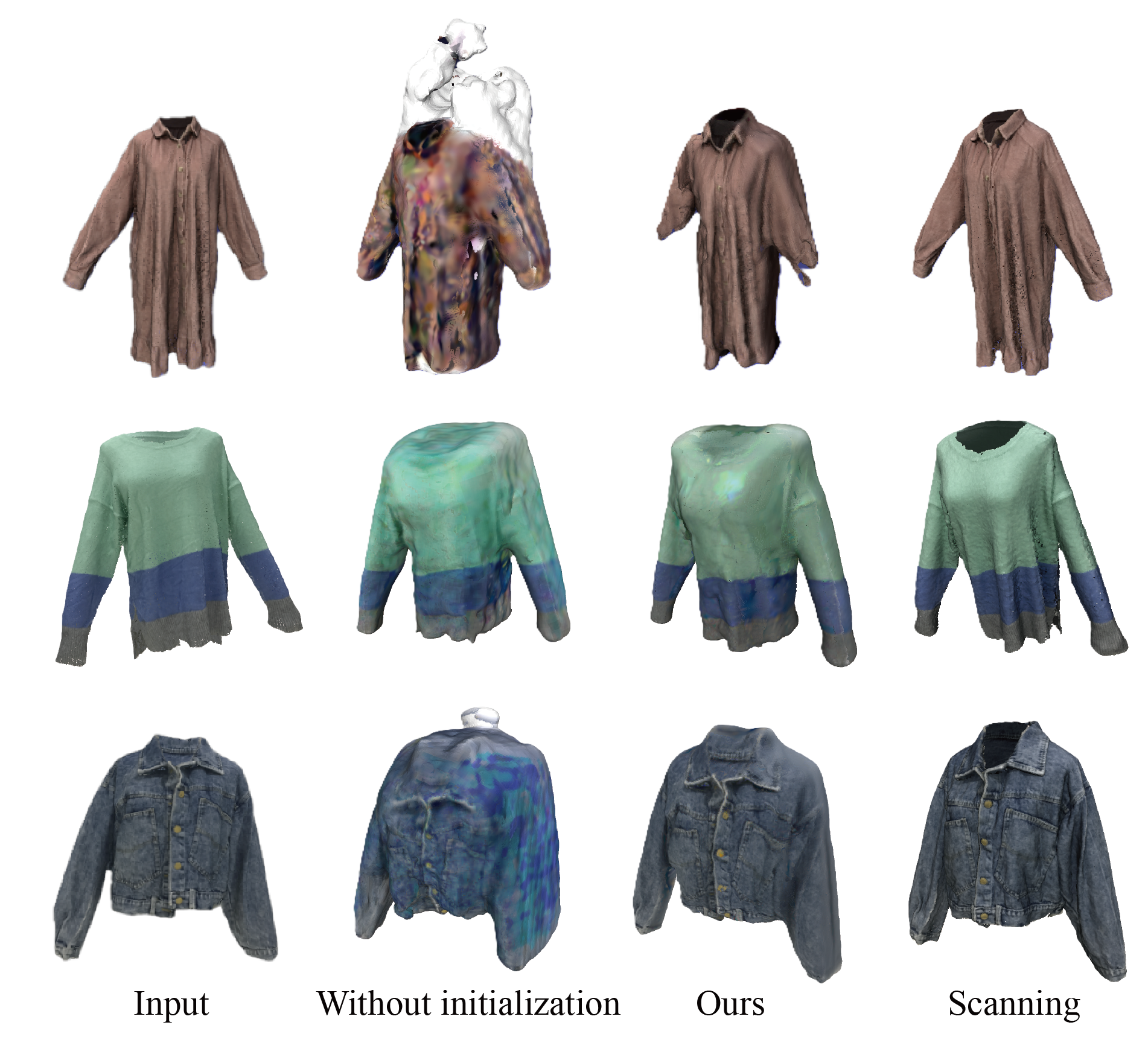}
   \caption{Ablation studies of the initialization with 3D Diffusion Model.}
   \label{fig:ablation}
 \end{figure}
In this ablation study, we compare the results of models with and without the initialization of 3D point clouds using 3D diffusion to verify that initialization can improve quality, as illustrated in Fig.\ref{fig:ablation}. 
The first column shows the input single-view image. The second column displays the results of optimizing randomly initialized 3D Gaussians using SDS loss. The third column showcases our method, which utilizes 3D diffusion to initialize the 3D Gaussians.
The model output without using 3D diffusion exhibits noticeable artifacts and noise, with distorted textures and unclear geometric structure of the garment. Additionally, 3D Gaussian points that should have been removed due to zero opacity remain intact.
In contrast, our method significantly enhances the 3D structure. The geometric structure of the 3D content is much clearer, the textures are more accurately represented, and there is less distortion and noise. Our method generates 3D assets that more closely approximate real scanned 3D assets.
The comparison clearly demonstrates the benefits of incorporating 3D diffusion in the initialization of point clouds. This module significantly enhances the overall quality of the generated 3D models by providing a more accurate and detailed representation of the geometric structure.

\subsubsection{Robustness to Varying Densities}
\begin{figure}[t]
   \centering
   \includegraphics[width=0.9\linewidth]{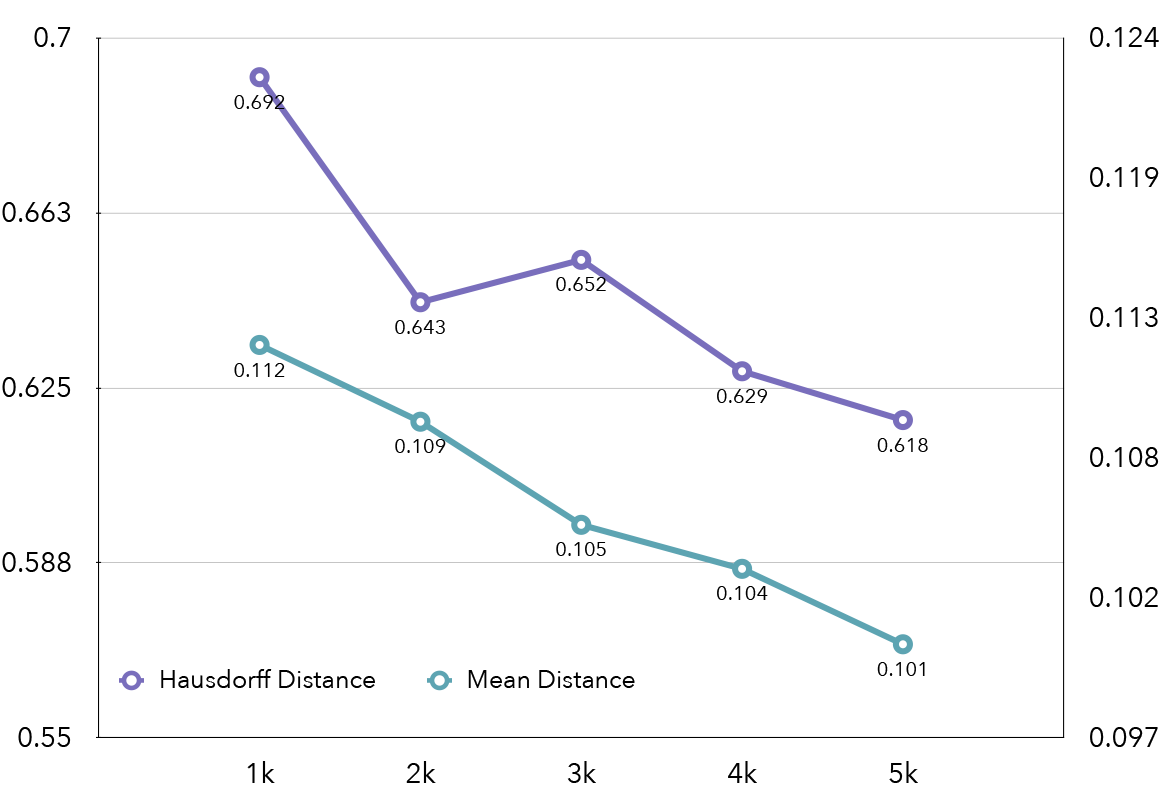}
   \caption{Hausdorff distance and Mean distance decrease when point clouds density increases. As the number of initial points continues to increase, the curve tends to decrease. The consistency of the initial point clouds impacts the quality of the optimization. Denser point clouds result in the optimized 3D content structure being closer to the reference scanned 3D structure.}
   \label{fig:density}
 \end{figure}

Fig.\ref{fig:density} demonstrates the impact of varying initial point cloud densities on the 3D content creation process, evaluated using two metrics: Hausdorff distance and Mean distance. 
As the point cloud density increases from 1k to 5k points, the Hausdorff distance significantly decreases from approximately 0.69 to about 0.418. The result indicates that the local structural error diminishes with increased point cloud density, suggesting that denser point clouds enhance the robustness of the local structure. Similarly, the Mean distance exhibits a downward trend, decreasing from around 0.112 to approximately 0.101. The consistent decline in Mean distance indicates a reduction in overall error, bringing the results closer to the actual target. This trend further validates the positive effect of higher point cloud densities on improving overall reconstruction accuracy.

\subsubsection{Scaling and Perturbation modules}
\begin{figure*}[t]
   \centering
   \includegraphics[width=\linewidth]{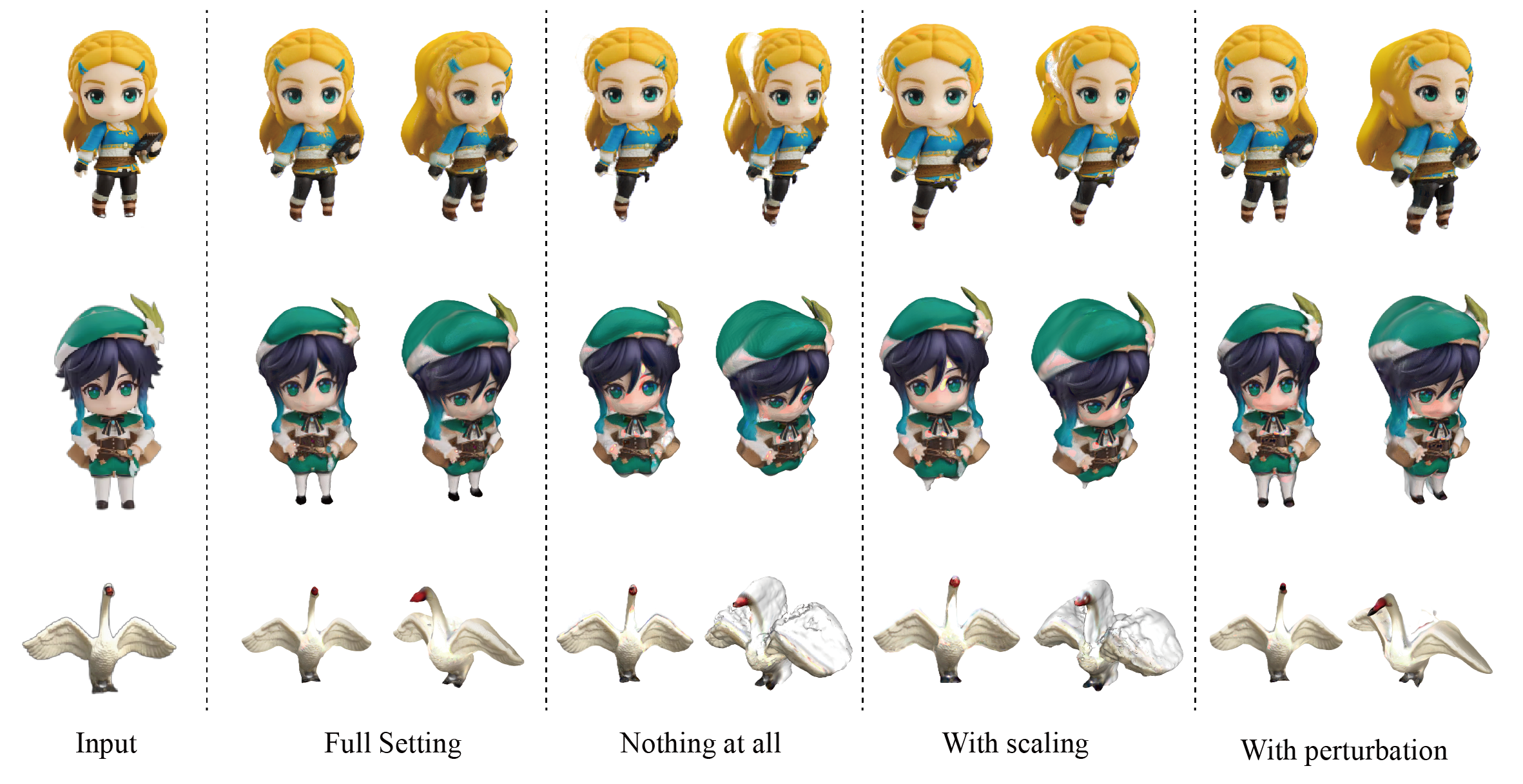}
   \caption{Ablation studies of Scaling and Perturbation.}
   \label{fig:ablation2}
 \end{figure*}

Based on the initial point clouds obtained from 3D diffusion, we also conduct an ablation study involving scaling and perturbation, as illustrated in Fig.\ref{fig:ablation2}. 
The first column shows the results with all modules included. The second column displays the results with only the 3D diffusion point cloud initialization, without any other modules. The third column presents the results with scaling but without noise. The fourth column shows the results with added noise but without scaling.
Comparing the second and third columns, adding the scaling module enhances the overall structure of the model. The 3D diffusion initialized point cloud is sparse in some areas (particularly sharp and fine parts) or blank regions, making it difficult to generate 3D Gaussians in these areas. Comparing the second and fourth columns, simply adding noise points to the 3D diffusion-generated point cloud significantly improves the issue of generating 3D Gaussians in blank regions.
According to the results in the first column, incorporating both the scaling and perturbation modules together provides a complementary effect. This not only improves the overall model performance, but also addresses the issue of generating 3D Gaussians in blank regions.

\subsubsection{Optimization Based on 2D Diffusion}
\begin{figure*}[t]
   \centering
   \includegraphics[width=\linewidth]{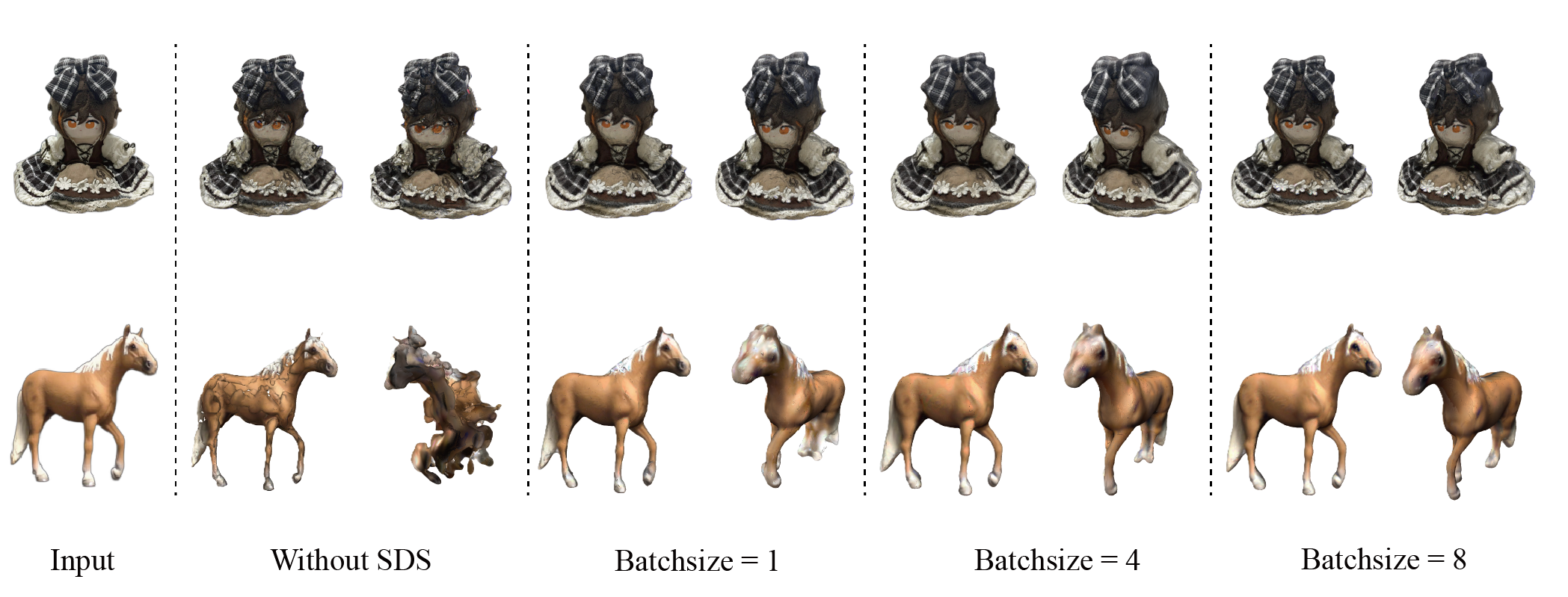}
   \caption{Ablation studies of 2D Diffusion with different batchsize.}
   \label{fig:ablation3}
 \end{figure*}
 
We present an ablation study examining the impact of different batch sizes in 2D diffusion on the generation of 3D models as Fig.\ref{fig:ablation3} shown. Models generated without using SDS (Score-Distance-Sampling) exhibit significant deficiencies in detail and overall structure. The textures are not refined, and the geometric shapes are inaccurate, indicating the importance of SDS in the generation process. With a batch size of 1, there is an improvement in the detail and structure of the model. However, the lack of sufficient diversity in the diffusion process leads to some artifacts and less clear textures. Increasing the batch size to 4 further improves the quality of the 3D content, providing better textures and geometric accuracy. This suggests a balance between the optimization of 3D Gaussians and diversity, thus enhancing the overall quality of the generated model. With a batch size of 8, the quality of the 3D assets improves further, but the generation time increases, reducing efficiency. Batch sizes of 4 and 8 provide significant improvements, with a batch size of 4 offering a good balance between computational efficiency and generation quality.

\section{Limitations and discussion}
Although we demonstrate the generalizability through experiments on DeepFashion3D, the 3D content generated from real-world images of dolls still exhibits blurred edges. Incorporating real depth information and optimizing the depth information rendered by 3D Gaussians could potentially improve the quality of the generated 3D assets. This represents a potential direction for future optimization.

Our framework incorporates 3D diffusion to initialize 3D Gaussians, significantly improving the structure of the generated 3D content and resulting in better geometric consistency. However, the texture of the generated 3D content is heavily influenced by 2D diffusion. Additionally, the 2D diffusion method we use is not designed for large-scale scenes, which limits the effectiveness of our approach in generating 3D assets for large-scale scenes.

\section{Conclusion}
In this work, we introduce a framework called Scaling Gaussian, capable of achieving a complete reconstruction of an object from a single image. 
We utilize 3D diffusion and 2D diffusion to optimize the representation of 3D Gaussian Splatting. 
Because 3D diffusion provides information of the prior point, it accelerates the convergence of 3D Gaussians during the optimization process.
Additionally, we introduce scaling and perturbation modules to further enhance the structural features of the 3D Gaussians and the generation efficiency. 
After converting to meshes, we further enhance details and sharpen edges using 2D diffusion, MSE loss, and GPP loss. 
This process allows our method to produce detailed and realistic geometric appearances with high geometric consistency. 
Our method demonstrated strong generalizability in experiments conducted on the DeepFashion3D dataset and other datasets.


\printcredits

\appendix 
\renewcommand{\thesection}{Appendix \Alph{section}}
\section{Smoothness of the L1 Loss}\label{appena}
The L1 loss function is defined as:
\begin{equation}
    \mathcal{L}_1(\theta) = \|y - f(\theta)\|_1 = \sum_{i=1}^n |y_i - f(\theta)_i|
\end{equation}
The L1 loss function is not differentiable at $y = f(\theta)$, rendering it non-smooth. In order to address this issue in practical applications,  Smooth L1 loss\cite{girshickICCV15fastrcnn} is frequently employed in order to guarantee smoothness.

\section{Smoothness of the Loss Function}\label{appen}
According to L-smooth, for any \(\theta\) and \(\theta'\), the gradient of \(\mathcal{L}_{r}\) satisfies the Lipschitz continuity condition:
\begin{equation}
    \|\mathcal{L}_{r}(\theta) - \mathcal{L}_{r}(\theta')\| \leq L \|\theta - \theta'\|
\end{equation}
The total loss function \(\mathcal{L}_{r}\) is:
\begin{equation}
\begin{aligned}
    \mathcal{L}_r(\theta) & = \lambda_{R} \| I^{r}_{R}(\theta) - \tilde{I}^{r}_{R} \|_2^2 + \lambda_{\alpha} \| I^{r}_{\alpha}(\theta) - \tilde{I}^{r}_{\alpha} \|_2^2 \\
   & +\lambda_{s} \mathbb{E}_{t,\Delta r,\epsilon} \| \epsilon_{\phi}(I_{R}^{r+\Delta r}(\theta); t, \tilde{I}_{R}^{r},\Delta r) - \epsilon \|_2^2 
\end{aligned}
\end{equation}
To demonstrate that it meets the smoothness conditions, we need to handle each term separately. We consider the Lipschitz continuity of each component.
\begin{equation}
\begin{aligned}
\| \lambda_{R} \| I^{r}_{R}(\theta) &- \tilde{I}^{r}_{R} \|_2^2 - \lambda_{R} \| I^{r}_{R}(\theta') - \tilde{I}^{r}_{R} \|_2^2 \| \\
& = \lambda_{R} \| \| I^{r}_{R}(\theta) - \tilde{I}^{r}_{R} \|_2^2 - \| I^{r}_{R}(\theta') - \tilde{I}^{r}_{R} \|_2^2 \| \\
& \leq \lambda_{R} L_{R} \| \theta - \theta' \|
\end{aligned}
\end{equation}
where \( L_{R} \) represents the Lipschitz constant of the function \( I^{r}_{R} \).
For the second term \( \lambda_{\alpha} \| I^{r}_{\alpha}(\theta) - \tilde{I}^{r}_{\alpha} \|_2^2 \):
\begin{equation}
\begin{aligned}
\| \lambda_{\alpha} \| I^{r}_{\alpha}(\theta) &- \tilde{I}^{r}_{\alpha} \|_2^2 - \lambda_{\alpha} \| I^{r}_{\alpha}(\theta') - \tilde{I}^{r}_{\alpha} \|_2^2 \| \\
& = \lambda_{\alpha} \| \| I^{r}_{\alpha}(\theta) - \tilde{I}^{r}_{\alpha} \|_2^2 - \| I^{r}_{\alpha}(\theta') - \tilde{I}^{r}_{\alpha} \|_2^2 \| \\
& \leq \lambda_{\alpha} L_{\alpha} \| \theta - \theta' \|
\end{aligned}    
\end{equation}
where \( L_{\alpha} \) represents the Lipschitz constant of the function \( I^{r}_{\alpha} \).
The third term, similar to the previous two terms, is an L2 loss and satisfies Lipschitz continuity($L_\phi$-smooth).
By combining these three terms, we obtain:
\begin{equation}
    \begin{aligned}
\| \mathcal{L}_r(\theta) - \mathcal{L}_r(\theta') \| \leq (\lambda_{R} L_{R} + \lambda_{\alpha} L_{\alpha} + \lambda_s L_\phi\ ) \| \theta - \theta' \|
\end{aligned}
\end{equation}
The optimization function \( \mathcal{L}_r(\theta) \) satisfies the smoothness condition, where \( L = \lambda_{R} L_{R} + \lambda_{\alpha} L_{\alpha} + \lambda_s L_\phi\ \).

\section*{Data availability}
The authors refer readers to the Experimental Setup section for details on how to download the datasets and code used in this work.

\section*{Acknowledgments}
This work was supported by East China University of Science and Technology.
\bibliographystyle{unsrt}
\bibliography{main}



\end{document}